\documentclass[sigconf]{acmart}

\usepackage{latexsym}
\usepackage{url}

\usepackage{xspace}
\usepackage{graphicx}
\usepackage{multirow}
\usepackage{makecell}
\usepackage{lscape}
\usepackage{bm}
\usepackage{algorithm}
\usepackage{algpseudocode}
\usepackage{framed}
\usepackage{color}
\usepackage{xcolor}
\usepackage{booktabs,colortbl}
\usepackage{bbm}
\usepackage{subfigure}
\usepackage{multirow}

\usepackage{amssymb}
\usepackage{amsmath}
\usepackage{soul}
\usepackage{multirow}
\usepackage{makecell}
\usepackage{tabularx}
\usepackage{booktabs}
\usepackage{setspace}
\usepackage{utfsym}
\usepackage{pifont}
\usepackage{arydshln}
\definecolor{lightgrayv}{HTML}{F4F3F8} 
\definecolor{redv}{HTML}{C00000}
\definecolor{bluev}{HTML}{0070C0}
\definecolor{grayv}{HTML}{707070}

\newcommand{\eg}{\emph{e.g.,}\xspace}
\newcommand{\ie}{\emph{i.e.,}\xspace}

\newcommand{\baby}{\textsc{Dm-inter}\xspace}

\AtBeginDocument{%
  \providecommand\BibTeX{{%
    \normalfont B\kern-0.5em{\scshape i\kern-0.25em b}\kern-0.8em\TeX}}}

\setcopyright{acmlicensed}
\copyrightyear{2024}
\acmYear{2024}

\acmConference[CIKM '24] {Proceedings of the 33rd ACM International Conference on Information and Knowledge Management}{October 21--25, 2024}{Boise, ID, USA.}
\acmBooktitle{Proceedings of the 33rd ACM International Conference on Information and Knowledge Management (CIKM '24), October 21--25, 2024, Boise, ID, USA}
\acmDOI{10.1145/3627673.3679799}
\acmISBN{979-8-4007-0436-9/24/10}

\begin{document}

\title{Why Misinformation is Created? Detecting them by Integrating Intent Features}

\author{Bing Wang}
\authornote{Key Laboratory of Symbolic Computation and Knowledge Engineering of the Ministry of Education, Jilin University.}
\orcid{0000-0002-1304-3718}
\affiliation{%
  \institution{College of Computer Science and Technology, Jilin University}
  \city{Changchun}
  \state{Jilin}
  \country{China}
}
\email{wangbing1416@gmail.com}

\author{Ximing Li}
\authornote{Corresponding author}
\authornotemark[1]
\orcid{0000-0001-8190-5087}
\affiliation{
  \institution{College of Computer Science and Technology, Jilin University}
  \city{Changchun}
  \state{Jilin}
  \country{China}
}
\email{liximing86@gmail.com}

\author{Changchun Li}
\authornotemark[1]
\orcid{0000-0002-8001-2655}
\affiliation{
  \institution{College of Computer Science and Technology, Jilin University}
  \city{Changchun}
  \state{Jilin}
  \country{China}
}
\email{changchunli93@gmail.com}

\author{Bo Fu}
\orcid{0000-0001-7030-821X}
\affiliation{
  \institution{Liaoning Normal University}
  \city{Dalian}
  \state{Liaoning}
  \country{China}
}
\email{fubocloud@163.com}

\author{Songwen Pei}
\orcid{0000-0003-0810-1458}
\affiliation{
  \institution{University of Shanghai for Science and Technology}
  \city{Shanghai}
  \country{China}
}
\email{swpei@usst.edu.cn}

\author{Shengsheng Wang}
\authornotemark[1]
\orcid{0000-0002-8503-8061}
\affiliation{
  \institution{College of Computer Science and Technology, Jilin University}
  \city{Changchun}
  \state{Jilin}
  \country{China}
}
\email{wss@jlu.edu.cn}

\renewcommand{\shortauthors}{Bing Wang et al.}

\begin{abstract}
  Various social media platforms, \eg Twitter and Reddit, allow people to disseminate a plethora of information more efficiently and conveniently. However, they are inevitably full of misinformation, causing damage to diverse aspects of our daily lives. To reduce the negative impact, timely identification of misinformation, namely \textbf{M}isinformation \textbf{D}etection (\textbf{MD}), has become an active research topic receiving widespread attention. As a complex phenomenon, the veracity of an article is influenced by various aspects. In this paper, we are inspired by the opposition of intents between misinformation and real information. Accordingly, we propose to reason the intent of articles and form the corresponding intent features to promote the veracity discrimination of article features. To achieve this, we build a hierarchy of a set of intents for both misinformation and real information by referring to the existing psychological theories, and we apply it to reason the intent of articles by progressively generating binary answers with an encoder-decoder structure. We form the corresponding intent features and integrate it with the token features to achieve more discriminative article features for MD. 
Upon these ideas, we suggest a novel MD method, namely \textbf{D}etecting \textbf{M}isinformation by \textbf{Inte}grating \textbf{Inte}nt featu\textbf{R}es (\textbf{\baby}). To evaluate the performance of \baby, we conduct extensive experiments on benchmark MD datasets. The experimental results validate that \baby can outperform the existing baseline MD methods. 

\end{abstract}



\maketitle

\begin{table}[t]
\centering
\renewcommand\arraystretch{1.15}
  \caption{Two claims about COVID-19 vaccine. They express different intents and have different veracity labels.}
  \label{example}
  \small
  \setlength{\tabcolsep}{5pt}{
  \begin{tabular}{m{7.9cm}}
    \toprule
    \textbf{Article}: Over-the-counter cold and cough medications are being pulled from drugstore shelves in an effort to start the ``next plandemic'' or force people to get the COVID-19 vaccine. \\
    \rowcolor{lightgrayv} \textbf{Veracity label}: {\color{bluev} \textit{Fake}}, \quad \textbf{Intent}: conspiracy theories \\
    \textbf{Source}: \url{https://apnews.com/article/fact-check-cold-cough-medicine-cvs-false-claims-860379163929} \\
    \hline
    \textbf{Article}: COVID-19 vaccines are safe for people who have existing health conditions, including conditions that have a higher risk of getting serious illness with COVID-19. \\
    \rowcolor{lightgrayv} \textbf{Veracity label}: {\color{redv} \textit{Real}}, \quad \textbf{Intent}: popularize commonsense \\
    \textbf{Source}: \url{https://www.mayoclinic.org/diseases-conditions/coronavirus/in-depth/coronavirus-vaccine/art-20484859} \\
    \bottomrule
  \end{tabular} }
\end{table}

\section{Introduction}

In our daily lives, various social media platforms, \eg Twitter, allow people to disseminate a plethora of information more efficiently and conveniently. However, these platforms are inevitably full of misinformation, causing damage to diverse aspects, \eg people's spirit and social economy \citep{messing2014selective,vosoughi2018the,van2022misinformation}. For example, a recent article demonstrated that a stagnant glass of water was seen in videos of Chinese astronauts, and its caption attempted to prove the footage wasn’t filmed in space.\footnote{\url{https://apnews.com/article/fact-check-misinformation-joe-biden-china-oceangate-9c470fb25fb34ead7e0f76864316e1a9}} Subsequent investigation confirmed this article as misinformation, and its goal is to smear China, and inflict significant damage on its international standing. 
To reduce the negative impact caused by misinformation, timely identification of them, namely \textbf{M}isinformation \textbf{D}etection (\textbf{MD}), has become an active research topic receiving widespread attention \citep{wu2023category,hu2023learn,huang2023faking}. Basically, the goal of MD is to automatically detect the veracity, \eg \textit{fake} and \textit{real}, of any article posted by social media users.

To handle the task of MD, many attempts have been made over the past decade \citep{zhang2021mining,nan2021mdfend,zhu2022generalizing,zhu2023memory}. Different from the traditional text classification problem, MD is much more challenging since the veracity of articles is influenced by diverse aspects, instead of the article content only. Therefore, the cutting-edge MD methods concentrate on extracting more discriminative features by incorporating influential aspects from psychology and sociology perspectives, including emotions \citep{shu2020early,zhang2021mining}, writing style \citep{wu2023category}, stance \citep{ma2018detect,yang2022a}, and propagation structure \citep{wei2021towards,yang2022a}. The recent MD studies further refer to auxiliary knowledge. For example, some MD methods extract entities from articles, and use their descriptions from the existing knowledge bases to promote the veracity discrimination of article features \citep{dun2021kan,hu2021compare,fung2021infosurgeon}.

Despite the effectiveness of those existing MD methods, this challenging task still suffers from a performance gap in practical applications due to its complexity. Inspired by the recent literature \citep{lu2022effects,wang2023understanding}, we observe that misinformation is often created by specific intents, which are often negative, and harmful. On the contrary, the real information tends to be more objective with the straightforward intent of sharing and popularizing. To clearly explain this inspiration, we illustrate two examples in Table~\ref{example}. Due to the opposition of intents between misinformation and real information, the intent of articles can be treated as an influential aspect in distinguishing the veracity of articles. 

\begin{figure}[t]
  \centering
  \includegraphics[scale=0.90]{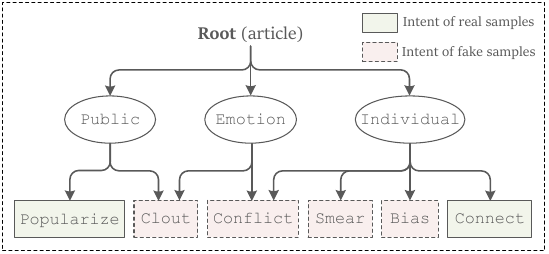}
  \caption{The hierarchy of expressed intents of articles.}
  \label{flow}
\end{figure}

Based on the aforementioned analysis, we propose to reason the intent of articles and form the corresponding intent features to promote the veracity discrimination of article features. To achieve this, we take inspiration from the tree-of-thought capability of language models \citep{yao2023tree}, and build a hierarchy of a set of intents for both misinformation and real information by referring to the existing psychological theories \citep{sikder2020minimalistic,lu2022effects} (as shown in Fig.~\ref{flow}). Given any article, we can reason its intent by progressively generating binary answers with the intent hierarchy. To be specific, in this paper, we employ the encoder-decoder structure backbone, \ie the T5 model \citep{raffel2020exploring}, where the encoder is used to extract token features of articles, and the decoder is used to reason the intent of articles and form the intent features. We can then integrate the token features with intent features to achieve more discriminative article features for MD. Upon these ideas, we suggest a novel MD method, namely \textbf{D}etecting \textbf{M}isinformation by \textbf{Inte}grating \textbf{Inte}nt featu\textbf{R}es (\textbf{\baby}). We evaluate \baby by conducting experiments on three benchmark MD datasets \textit{GossipCop} \citep{shu2020fakenewsnet}, \textit{PolitiFact} \citep{popat2018declare}, and \textit{Snopes} \citep{popat2017where}. The main experimental results can indicate that \baby can consistently improve the performance of any MD baseline models. Additionally, we also provide some cases to demonstrate our model can explicitly reason intents. 

The contributions of this paper can be summarized as follows:
\begin{itemize}
    \item We present that the intent of articles is an influential aspect of MD and propose a new MD method, named \baby, to integrate intent features into misinformation detectors.
    \item To implement the reasoning of intents, we design an encoder-decoder structure and specify its decoder side to progressively reason intents and form intent features.
    \item Extensive experiments are conducted to indicate the effectiveness of \baby in improving the performance of baseline MD methods.
\end{itemize}

\section{Related Works}

We systematically review the previous published works on MD and intent reasoning, and briefly summarize them in this section.

\subsection{Misinformation Detection}

Misinformation, based on the data format of the information, can be classified as fake news and rumors, which has caused seriously negative impacts on the economy, politics, and more \citep{vosoughi2018the,van2022misinformation}. To alleviate their effects, MD is emergent as an active topic. Existing methods for detecting misinformation primarily focus on content-based approaches \citep{shu2020fakenewsnet,zhang2021mining,wu2023category,hu2023learn,huang2023faking,wang2024escaping}. In general, they employ a variety of deep learning models to learn the potential correlations between content and veracity labels. 
In addition to this, content-based MD methods also introduce various external features to aid detection \citep{zhang2021mining,zhu2022generalizing,zhu2023memory}. For example, some works propose to introduce domain labels to enhance the performance of MD models across different domains, thereby improving its overall performance on the dataset \citep{nan2021mdfend,nan2022improving,zhu2023memory}. 
Beyond these discriminative approaches, with the rapid development of large language models, some arts have harnessed the world knowledge stored within these models, and prompt them to generate veracity predictions and even their underlying rationales in a generative manner \citep{hu2024bad,wan2024dell,wang2024explainable}.

Unlike these approaches, we find that the intent of articles also determines the veracity of the news. For instance, if an article exhibits clear racial bias, it is more likely to be misinformation. To this end, we propose to progressively reason the potential intent behind the article and mine the intent features to enhance MD methods.
Within the community, there are few efforts that explore the relationship between intent and misinformation. These studies primarily analyze the intents of misinformation spreaders on social media \citep{zhou2022this,agarwal2023understanding}, instead of focusing on the potential role of misinformation creators for the MD task. For example, \citet{agarwal2023understanding} heuristically assess the intent of fake news spreaders by presenting an influence graph; \citet{lee2021intention,guo2023detecting,wang2023understanding} detect the intents of fake news articles, but neglect the effect of intents on predicting veracities.

\begin{figure*}[t]
  \centering
  \includegraphics[scale=1.30]{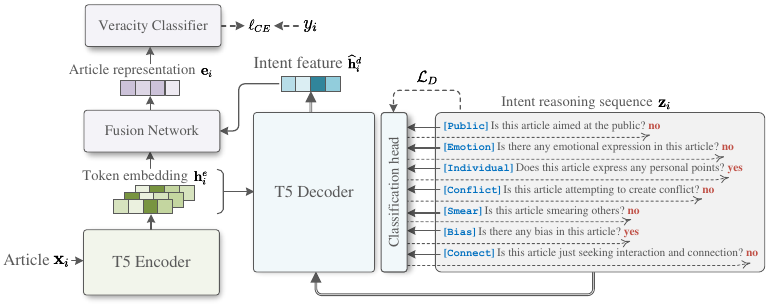}
  \caption{The overall framework of \textbf{\baby}. An article $\mathbf{x}_i$ is fed into the T5 encoder to obtain an article feature $\mathbf{h}_i^e$. This feature is then input into the T5 decoder to reason the intents and extract the intent feature $\mathbf{\widehat h}_i^d$. Finally, the two features are fused and classified for veracity.}
  \label{framework}
\end{figure*}

\subsection{Intent Reasoning}

\vspace{3pt}
\noindent
\textbf{Intent detection}.
This work focuses on uncovering the intents behind the misinformation creators. 
In the natural language processing community, several studies have delved into detecting the intent expressed by textual content \citep{larson2019an,zhou2022knn,firdaus2023multitask}. 
This task can be employed to enable conversational agents to identify potential purposes in user dialogues and provide responses promptly. 
Current studies primarily concentrate on the typical supervised learning settings, where the data distribution and corresponding intent labels for the training and testing sets are identical \citep{zhang2021deep,firdaus2023multitask}, or in a more practical scenario, involving the detection of out-of-distribution intents \citep{larson2019an,zhou2022knn,zhou2023two}. These efforts aim to train prominent intent detection models under the condition of existing out-of-distribution data in the training set.
In this paper, due to the lack of ground-truth intent labels in MD datasets, our focus is on a more challenging scenario, namely zero-shot intent detection. Unlike zero-shot tasks where an additional support set is provided for training, and then the trained model is transferred to new data \citep{siddique2021generalized,liu2022a}, our scenario involves identifying intents without any supervised data. To address this challenge, recent efforts leverage the reasoning capabilities of pre-trained language models (PLMs) \citep{brown2020language,wei2022chain}. Through the use of manually crafted prompt queries, various prompting strategies are designed for reasoning \citep{kuo2023zero,zhang2023revisit,zhang2024intention}.

\vspace{3pt}
\noindent
\textbf{Reasoning on language models}.
Recently, with the rapid development of cutting-edge pre-training techniques, a considerable amount of arts leverage PLMs for reasoning in various areas such as sentiment analysis \citep{fei2023reasoning,ouyang2024aspect}, mathematics \citep{wei2022chain,zhang2023automatic}, and multi-modality \citep{zhang2023multimodal}. Beyond directly prompting language models to generate answers, some studies propose using PLMs for chained reasoning, \eg Chain-of-Thought (CoT) \citep{wei2022chain,kojima2022large,zhang2023automatic} and tree-of-thought \citep{yao2023tree}, to explicitly generate the reasoning process. Accordingly, our work draws inspiration from these approaches to manually construct an intent hierarchy and utilize PLMs for progressive reasoning within this hierarchical structure.

\section{Our Proposed Method}

In this section, we briefly describe the task formulation of MD, and then introduce the proposed \baby method in detail.

\vspace{3pt}
\noindent
\textbf{Problem formulation of MD.}
Formally, each training sample $(\mathbf{x}, y)$ is composed of a raw article $\mathbf{x}$ and its ground-truth veracity label $y \in \{0, 1\}$, where $y = 0 / 1$ represents the article is real / fake. Given a collection of $N$ training samples $\mathcal{D} = \{(\mathbf{x}_i, y_i)\}_{i=1}^N$, the goal of MD is to train a misinformation detector $\mathcal{F}_{\Theta}(\cdot)$ with $\mathcal{D}$ to predict veracity labels for unseen articles. For clarity, the important notations and their descriptions are summarized in Table~\ref{notation}.

\begin{table}[t]
\centering
\renewcommand\arraystretch{1.15}
\small
  \caption{Summary of notations.}
  \label{notation}
  \setlength{\tabcolsep}{5pt}{
  \begin{tabular}{m{1.6cm}<{\centering}|m{6.0cm}<{\centering}}
    \bottomrule
    Notation & Description \\
    \hline
    $N$ & number of training samples \\
    $\mathbf{x}_i$, $y_i$ & training news article and its veracity label \\
    $\Phi, \Pi, \Psi, \mathbf{W}$ & parameters of the misinformation detector \\
    $\mathbf{h}_i^e$, $\mathbf{h}_i^d$ & token embeddings from the encoder and decoder \\
    $\mathbf{\widehat h}_i^e$ & intent feature \\
    $\mathbf{e}_i$ & article representation \\
    $\alpha_i$ & adaptive sample weight \\
    $T_i$ & number of reasoning steps \\
    $\mathbf{Q}_{it}$ & intent query at the $t$-th reasoning step \\
    $a_{it}$ & generated answer at the $t$-th reasoning step \\
    \bottomrule
  \end{tabular} }
\end{table}

\subsection{Overview of \baby}

As illustrated in Fig.~\ref{framework}, \baby consists of the following four components. (1) \textit{LM encoder}, specified by the T5 encoder \citep{raffel2020exploring}, extracts the hidden token embeddings of a given article. (2) \textit{LM decoder}, initialized by the T5 decoder, reasons intents on the pre-defined intent hierarchy and generate the intent features. (3) \textit{Fusion network} integrates the token embeddings and intent features into an overall article representation. Finally, (4) \textit{veracity classifier} provides the veracity prediction of the article.

Formally, given an article $\mathbf{x}_i = \{x_{ij}\}_{j=1}^L$, where $L$ indicates the length of the article, we use a pre-trained T5 encoder $\mathcal{F}_{\Phi}(\cdot)$ to generate its hidden token embeddings $\mathbf{h}_i^e = \mathcal{F}_{\Phi}(\mathbf{x}_i) = \{\mathbf{h}_{ij}^e\}_{j=1}^L$. 
Then, to reason the potential intent behind $\mathbf{x}_i$, we refer to some psychological concepts \citep{sikder2020minimalistic,lu2022effects} and present an \textbf{intent hierarchy} in Fig.~\ref{flow}, which can formulate intent detection as a hierarchical text classification problem \citep{zhou2020hierarchy,wang2022incorporating} along the hierarchy, and we solve it by using a generative approach. Specifically, for intents in the pre-defined set $\mathcal{I} = $ \{ \texttt{Public}, \texttt{Emotion}, \texttt{Individual}, \texttt{Popularize}, \texttt{Clout}, \texttt{Conflict}, \texttt{Smear}, \texttt{Bias}, \texttt{Connect} \}, we manually design their corresponding natural language queries $\mathcal{Q} = \{\mathbf{Q}_k\}_{k=1}^{|\mathcal{I}|}$, which can be answered by \textit{yes} / \textit{no} (specific implementation is described in Sec.~\ref{sec3.2}). 
Based on these queries, we progressively prompt a T5 decoder $\mathcal{F}_\Pi(\cdot)$ to reason one or multiple paths on the hierarchy, and obtain a textual \textbf{intent reasoning sequence}, which is represented as $\mathbf{z}_i = \{z_{ij}\}_{j=1}^{\widetilde L}$, where $\widetilde L$ denotes the length of the sequence.
To further supervise the decoder to generate answers for unseen articles, we introduce an auto-regressive objective $\mathcal{L}_{D}$ with $\mathbf{z}_i$ to conduct a self-training strategy.
Meanwhile, during the reasoning process, the T5 decoder also outputs the token embeddings of $\mathbf{z}_i$, represented by $\mathbf{h}_i^d = \mathcal{F}_\Pi(\mathbf{z}_i) = \{\mathbf{h}_{ij}^d\}_{j=1}^{\widetilde L}$. Then, we directly adopt the following average pooled representation, which can be seen as the \textbf{intent feature}.
\begin{equation}
    \label{eq2}
    \mathbf{\widehat h}_i^d = \frac{1}{\widetilde L} \sum \nolimits _{j=1}^{\widetilde L} \mathbf{h}_{ij}^d.
\end{equation}

Accordingly, given two features $\mathbf{h}_i^e$ and $\mathbf{\widehat h}_i^d$, we adopt a multi-head attention network to fuse them to obtain the overall article representation $\mathbf{e}_i = \mathcal{F}_{\Psi}(\mathbf{h}_i^e, \mathbf{\widehat h}_i^d)$, and formulate the following objective to optimize the model parameters.
\begin{equation}
    \label{eq3}
    \mathop{\boldsymbol{\min}} \limits _{\Phi, \Pi, \Psi, \mathbf{W}} \mathcal{L} = \frac{1}{N} \sum \nolimits _{i=1}^N \alpha_i \left( \ell_{CE} \big( \mathbf{e}_i \mathbf{W}, y_i \big) + \beta \mathcal{L}_{D} \right),
\end{equation}
where $\mathbf{W}$ represents a learnable veracity classifier, $\ell_{CE}(\cdot, \cdot)$ denotes a cross-entropy loss function, and $\beta$ is a trade-off parameter to balance two objectives, respectively. 
Additionally, we further improve \baby by heuristically assigning weights $\alpha_i$ for training samples to alleviate the error propagation and veracity inconsistency problems.
In the following sections, we will introduce the details of intent feature extraction, veracity feature fusion, and adaptive weight assigning, respectively.

\subsection{Intent Feature Extraction and Optimization} \label{sec3.2}
Given an article $\mathbf{x}_i$, we extract its intent reasoning sequence $\mathbf{z}_i$ and intent feature $\mathbf{\widehat h}_i^d$ with an LM decoder $\mathcal{F}_{\Phi}(\cdot)$, and optimize it by using a self-training strategy. To achieve this, we follow the implementation process below.

\vspace{3pt}
\noindent
\textbf{Intent hierarchy and queries.}
We review the viewpoints expressed by the psychological literature \citep{vosoughi2018the,scheufele2019science,lu2022effects} on the question, ``\textit{why misinformation is created?}''. The answers can be referred to as \textit{intents}. Based on their granularity, we build an intent hierarchy, shown in Fig.~\ref{flow}, enabling us to reason intents from coarse to fine granularity (from top to bottom on the hierarchy) and thereby improving the reliability of intent reasoning. Specifically, these pre-defined intents and their corresponding descriptions are detailed as follows:
\begin{itemize}
    \item \texttt{Public}. The article intends to influence public perceptions.
    \item \texttt{Emotion} \citep{weeks2015emotions,vosoughi2018the,zhang2021mining}. The article aims to express emotions or stir up emotions in others.
    \item \texttt{Individual}. The article conveys opinions about specific individuals or events.
    \item \texttt{Popularize} \citep{umbricht2016push}. The article seeks to disseminate knowledge, \eg scientific facts and current societal issues, thereby providing valuable information to enhance public awareness of ongoing events and problems.
    \item \texttt{Clout} \citep{lu2022effects}. The article is strategically crafted to attract the attention of other social media users to achieve profit. These articles often feature attention-grabbing headlines and content.
    \item \texttt{Conflict} \citep{mazepus2023information}. The article attempts to create conflicts between certain individuals or groups for political purposes or other improper reasons.
    \item \texttt{Smear} \citep{sawano2019combating}. The article deliberately smears and attacks specific individuals or organizations, due to personal animosity, political opposition, or competitive relationships.
    \item \texttt{Bias} \citep{sikder2020minimalistic}. The article expresses an explicit personal bias.
    \item \texttt{Connect}. The article provides a platform for individuals to share authentic perspectives on the same event or topic, fostering social connections and common interests.
    \item \texttt{NoIntent} \citep{scheufele2019science,zhou2022this}. The intent expressed by the article is not among the intents listed above, or the article does not express a clear intent, \eg simply sharing.
\end{itemize}

Afterwards, to prompt an LM decoder to reason intents, we design several natural language queries for the above intents, which is denoted as $\mathcal{Q} = \{\mathbf{Q}_k\}_{k=1}^{|\mathcal{I}|}$, in Table~\ref{prompt}.

\begin{table}[t]
\centering
\renewcommand\arraystretch{1.10}
  \caption{Designed prompts for intent reasoning.}
  \label{prompt}
  \small
  \setlength{\tabcolsep}{5pt}{
  \begin{tabular}{m{1.6cm}<{\centering}|m{6.2cm}<{\centering}}
    \bottomrule
    query & prompts \\
    \hline
    \texttt{Public} & Is this article aimed at the public? \\
    \texttt{Emotion} & Is there any emotional expression in this article? \\
    \texttt{Individual} & Does this article express any personal points? \\
    \texttt{Popularize} & Is this an article aimed at popularization? \\
    \texttt{Clout} & Is this an article aimed at pursuing attention? \\
    \texttt{Conflict} & Is this article attempting to create conflict? \\
    \texttt{Smear} & Is this article smearing others? \\
    \texttt{Bias} & Is there any bias in this article? \\
    \texttt{Connect} & Is this article just seeking interaction and connection? \\
    \bottomrule
  \end{tabular} }
\end{table}

\vspace{3pt}
\noindent
\textbf{Intent reasoning and feature extraction.}
Based on the above intent hierarchy and queries, we can conduct reasoning of intents and extract their features. Specifically, following a breadth-first search strategy, we first prompt the LM decoder with corresponding queries for intents at the second layer of the hierarchy, \eg \texttt{Emotion}, and generate binary responses of \textit{yes} / \textit{no}. Then, we proceed to query the intents at the third layer. If the answer to the parent intent of one intent is \textit{yes}, we query it; otherwise, it will be skipped. Through this search strategy, we get a \textbf{reasoning path} with $T_i$ reasoning steps for $\mathbf{x}_i$. By progressively answering the queries on this reasoning path, we can obtain a text sequence $\mathbf{z}_i$ to represent the intent context and extract its feature as the intent feature $\mathbf{h}_i^d$.
Formally, to generate an answer $a_{it}$ corresponding to current reasoning step $t$, given each article $\mathbf{x}_i$, we progressively input its hidden representation $\mathbf{h}_i^e$ from the LM encoder and an intent query $\mathbf{Q}_{it}$ of the reasoning step $t$ into the LM decoder, and generate the hidden token representations $\mathbf{h}_{it}^d$ as follows:
\begin{equation}
    \label{eq4}
    \mathbf{h}_{it}^d = \mathcal{F}_{\Pi} \left( \mathbf{h}_i^e, \mathcal{A}_{i, t-1} \oplus \mathbf{Q}_{it} \right), \quad t = \{1, 2, \cdots, T_i\},
\end{equation}
where $\mathcal{A}_{i, t-1}$ denotes the cumulative answer sequence of the previous reasoning step $t-1$, and $\oplus$ represents a concatenation operation.
Then, we input $\mathbf{h}_{it}^d$ into a vocabulary-level classification head to obtain a logit $\mathbf{q}_{it} = \mathbf{h}_{it}^d \mathbf{W}_D$, which indicates the probabilities of predicted tokens. We classify the answer $a_{it} \in \{\textit{yes}, \textit{no}\}$ corresponding to the query $\mathbf{Q}_{it}$ by checking the predicted logit $q_{it}$ of the last token in $\mathbf{q}_{it}$, $a_{it} = \textit{yes}$ if the probability of predicting \textit{yes} is greater than the probability of predicting \textit{no}; otherwise, $a_{it} = \textit{no}$. Afterwards, we update the answer sequence $\mathcal{A}_{it}$ for the next reasoning path as follows:
\begin{equation}
    \label{eq5}
    \mathcal{A}_{it} = \mathcal{A}_{i, t-1} \oplus \mathbf{Q}_{it} \oplus a_{it}, \quad t = \{1, 2, \cdots, T_i\}.
\end{equation}
When all $T_i$ intents in the reasoning path have been reasoned, $\mathcal{A}_{i{T_i}}$ and $\mathbf{h}_{i{T_i}}^d$ at the final time step ${T_i}$ serves as $\mathbf{z}_i$ and $\mathbf{h}_{i}^d$, respectively. Importantly, if all answers $\{a_{it}\}_{t=1}^{T_i}$ throughout the reasoning path are \textit{no}, it indicates that the article $\mathbf{x}_i$ merely share information without expressing any explicit intents, or the expressed intents fall outside the defined intent hierarchy. In this scenario, a sentence ``\textit{This article does not convey any intents}'' will be appended to $\mathbf{z}_i$.

\vspace{3pt}
\noindent
\textbf{Self-training optimization.}
To enable the decoder to generate more accurate answers for unseen articles, we involve training it with a specific objective. Unfortunately, the ground-truth intent label of $\mathbf{x}_i$ is unobserved. Therefore, we take inspiration from the self-training strategy in the weakly-supervised community \citep{zou2019confidence,li2021semi} to design the objective. Typically, the self-training method supervises one sample with its own sharpened prediction. In this work, the sequence $\mathbf{z}_i = \{z_{ij}\}_{j=1}^{\widetilde L}$ is the naturally sharpened label, and we use it to supervise our MD model with an auto-regressive objective as follows:
\begin{equation}
    \label{eq6}
    \mathcal{L}_{D} = \frac{1}{N{\widetilde L}} \sum \nolimits _{i=1}^N \sum \nolimits _{j=1}^{\widetilde L} \ell_{CE} \left( z_{ij}, P(o_{ij} | \mathbf{o}_{i,<j}, \mathbf{h}_i^e; \Pi) \right),
\end{equation}
where $\mathbf{o}_i = \{o_{ij}\}_{j=1}^{\widetilde L}$ indicates the generated textual sequence corresponding to $\mathbf{x}_i$ from the LM decoder, and $P(o_{ij} | \cdot\ ; \Pi)$ represents the output probability of $o_{ij}$ from the LM decoder.

\subsection{Veracity Feature Fusion} \label{sec3.3}

Given the token embedding $\mathbf{h}_i^e$ output from the encoder and the intent feature $\mathbf{\widehat h}_i^d$, we fuse them into an overall article representation $\mathbf{e}_i$ and use it for veracity prediction. Specifically, we leverage a confidence-guided attention network to fuse them as follows: 
\begin{equation} 
    \label{eq7}
    \mathbf{e}_i = \text{Att} \left( c_i \mathbf{\widehat h}_i^d \mathbf{W}_Q, \mathbf{h}_i^e \mathbf{W}_K, \mathbf{h}_i^e \mathbf{W}_V \right),
\end{equation}
where $\text{Att}(\cdot)$ is a typical attention network \citep{vaswani2017attention}, and $\mathbf{W}_Q$, $\mathbf{W}_K$ and $\mathbf{W}_V$ are learnable query, key and value parameters, respectively. Additionally, we take into account that the confidence of intent answers $\{a_{it}\}_{t=1}^{T_i}$ can impact the credibility of intent features. If the output confidences of answers are low, it is advisable to assign a smaller weight to the corresponding intent feature $\mathbf{\widehat h}_i^d$ during the feature fusion stage. Building upon this consideration, we propose to assign an adaptive weight $c_i$ to intent features $\mathbf{\widehat h}_i^d$, calculated as the average probability value of the $T_i$ answers generated during the intent reasoning process.

\renewcommand{\algorithmicrequire}{\textbf{Input:}}
\renewcommand{\algorithmicensure}{\textbf{Output:}}
\begin{algorithm}[t]
\setstretch{1.0}
    \caption{Overall training process of \baby.}
    \label{algorithm}
    \begin{algorithmic}[1]
    \Require Training dataset $\mathcal{D} = \{\mathbf{x}_i, y_i\}_{i=1}^N$; trade-off parameters $\beta$; training iteration number $\Gamma$; pre-defined intent hierarchy and its queries $\{\mathbf{Q}_k\}_{k=1}^{|\mathcal{I}|}$; reasoning steps $\{T_i\}_{i=1}^N$.
    \Ensure Model parameters $\Theta = \{\Phi, \Pi, \Psi, \mathbf{W}\}$; intent reasoning sequence $\{\mathbf{z}_i\}_{i=1}^N$.
    \State Initialize $\{\Phi, \Pi\}$ by the pre-trained T5 model, and other parameters in $\mathbf{\Theta}$ randomly.
    \For{$\gamma = 1, 2, \cdots, \Gamma$} 
    \State Draw a mini-batch $\mathcal{B}$ from $\mathcal{D}$ randomly.
    \State Get $\mathbf{h}^e$ with the encoder $\mathcal{F}_{\Phi}(\cdot)$.
    \State Reason intent on the intent hierarchy following Sec.~\ref{sec3.2}, and obtain $\mathbf{z}$, $\mathbf{\widehat h}^d$, and calculate $\mathcal{L}_D$ using Eq.~\eqref{eq6}.
    \State Fuse $\mathbf{h}^e$ and $\mathbf{\widehat h}^d$ using Eq.~\eqref{eq7}.
    \State Calculate weight $\alpha$ following Sec.~\ref{sec3.4}.
    \State Optimize $\Theta$ with $\mathcal{L}$ in Eq.~\eqref{eq3}.
    \EndFor
    \end{algorithmic}
\end{algorithm}

\subsection{Adaptive Weight Assigning} \label{sec3.4}

During the reasoning on the intent hierarchy, two challenges inevitably arise: \textbf{error propagation} and \textbf{veracity inconsistency}. To alleviate these issues, we assign an adaptive weight, denoted as $\alpha_i$, in Eq.~\eqref{eq3} for each training sample. The specific descriptions are as follows:

\vspace{3pt}
\noindent
\textbf{Error propagation}.
The error propagation issue means that if an intent in the intent hierarchy is reasoned incorrectly, then its child intents will also be incorrect. To mitigate this, we design a weight $\alpha_i^E$. Specifically, for the reasoning reasoning steps $\{1, 2, \cdots, T_i\}$ described in Sec.~\ref{sec3.2}, we reversely re-reason them and output new answers $\{\widehat a_{it}\}_{t={T_1}}^1$. If there is a significant difference between the original and reversed answers, indicating that the error propagation exists, and we will assign a lower weight to this sample. We implement the calculation of the weight with an L2 norm as follows:
\begin{equation}
    \label{eq8}
    \alpha_i^E = \frac{T}{\sum \nolimits_{t=1}^T \|a_{it} - {\widehat a}_{it}\|_2^2}.
\end{equation}

\vspace{3pt}
\noindent
\textbf{Veracity inconsistency}.
Given the pre-defined hierarchy shown in Fig.~\ref{flow}, each intent corresponds to a veracity label, \eg the article expressing \texttt{Bias} tends to be \textit{fake}. Based on this premise, we suggest that when the reasoned intent of an article fails to align with its veracity label, indicating an incorrect reasoning for this sample, a low weight should be assigned. Therefore, we achieve it with the objective as follows:
\begin{equation}
    \label{eq9}
    \alpha_i^V = \left\{
	\begin{aligned}
	       \  1,& \quad \text{veracity consistency}, \\
	       \  0,& \quad \text{veracity inconsistency}. \\
	\end{aligned}
	\right .
\end{equation}

Finally, we average these two weights to obtain the final sample weight $\alpha_i = \big( \alpha_i^E + \alpha_i^V \big) \big/ 2$ for training in Eq.~\eqref{eq3}, and the overall training process is presented in Alg.~\ref{algorithm}.

\section{Experiments}

In this section, we conduct extensive experiments for answering the following questions:
\begin{itemize}
    \item \textbf{Q1}: How effective is our proposed model \baby in integrating intent features compared to the backbone models without it?
    \item \textbf{Q2}: Does each module in \baby have a promoting effect on the overall performance?
    \item \textbf{Q3}: Can our proposed \baby accurately and explicitly reason intents of articles?
\end{itemize}

\subsection{Experimental Settings}

\vspace{3pt}
\noindent
\textbf{Datasets}.
We conduct our experiments across three benchmark MD datasets to evaluate the performance of \baby. For clarity, their statistics are demonstrated in Table~\ref{datasetsta}, and we declare their details as follows:

\begin{itemize}
    \item \textit{GossipCop} is a prevalent MD dataset collected by \citet{shu2020fakenewsnet}, which totally includes 11,800 pairs of articles and their veracity labels. We follow \citet{zhu2022generalizing} to divide \textit{GossipCop} into training, validation, and test subsets. Specifically, they divide the dataset according to the time stamps of articles, which is a more practical and challenging setting. Its training articles were posted between 2000 and 2017, and test and validation samples were published in 2018, respectively.
    \item \textit{PolitiFact} is provided by \citet{popat2018declare}. Its original veracity labels include \{\textit{true}, \textit{mostly true}, \textit{half true}, \textit{mostly false}, \textit{false}, \textit{pants-on-fire}\}. Following \citet{rashkin2017truth}, we incorporate \{\textit{true}, \textit{mostly true}, \textit{half true}\} into the \textit{real} class, and \{\textit{mostly false}, \textit{false}, \textit{pants-on-fire}\} into \textit{fake}. Additionally, we divide the dataset by following the settings of \citet{vo2021hierarchical}.
    \item \textit{Snopes}\footnote{You can download the complete versions of \textit{PolitiFact} and \textit{Snopes} from \url{https://www.mpi-inf.mpg.de/dl-cred-analysis/}.} \citep{popat2017where} is collected from a fact-checking website, and the veracity labels have been manually verified by human experts, and we follow \citet{vo2021hierarchical} to divide this dataset.
\end{itemize}

\vspace{3pt}
\noindent
\textbf{Baselines}.
We thoroughly compare our \baby to several baseline MD models to demonstrate its performance. They are briefly described as follows:

\begin{itemize}
    \item BERT \citep{devlin2019bert} is a prevalent language model pre-trained using masked tokens and next-sentence prediction objectives. It is typically employed to extract semantic features from texts.
    \item T5 \citep{raffel2020exploring} is the language model pre-trained for the text generation task. It consists of an encoder to capture features of input queries and a decoder to auto-regressively generate target answers, respectively. For the MD task, we implement it by inputting an article into both the encoder and decoder sides and fusing their embeddings output from the encoder and decoder with the attention network described in Sec.~\ref{sec3.3}.
    \item EANN \citep{wang2018eann} concentrates on detecting multimodal misinformation in emergent events, which learns event-invariant news features with an adversarial strategy. In this work, we follow the implementation of \citet{zhu2022generalizing}, wherein the image modality is omitted, and event labels are directly assigned based on article timestamps.
    \item BERT-EMO \citep{zhang2021mining} introduces additional emotional signals to enhance MD models. They extract the emotional features of articles by utilizing existing emotion dictionaries.
    \item CED \citep{wu2023category} is also an encoder-decoder structure designed for MD, which leverages the encoder to extract news semantic representations and the decoder to generate intra/inter-category features. It is the current state-of-the-art MD model.
\end{itemize}

We re-produce all these baseline models, and implement EANN, BERT-EMO and CED by separately using BERT, T5, and our proposed model \baby as backbones, resulting in some combined versions, \eg T5 + EANN, EANN + \baby.

\begin{table}[t]
\centering
\renewcommand\arraystretch{1.10}
  \caption{Statistics of three prevalent MD datasets.}
  \label{datasetsta}
  \setlength{\tabcolsep}{5pt}{
  \begin{tabular}{m{1.80cm}<{\centering}m{0.62cm}<{\centering}m{0.62cm}<{\centering}m{0.62cm}<{\centering}m{0.62cm}<{\centering}m{0.62cm}<{\centering}m{0.62cm}<{\centering}}
    \toprule
    \multirow{2}{*}{Dataset} & \multicolumn{2}{c}{\# Train} & \multicolumn{2}{c}{\# Validation} & \multicolumn{2}{c}{\# Test} \\
    \cmidrule(r){2-3} \cmidrule(r){4-5} \cmidrule(r){6-7}
    & Fake & Real & Fake & Real & Fake & Real \\
    \hline
    \textit{GossipCop} & 2,024 & 5,039 & 604 & 1,774 & 601 & 1,758 \\
    \textit{PolitiFact} & 1,224 & 1,344 & 170 & 186 & 307 & 337 \\
    \textit{Snopes} & 2,288 & 838 & 317 & 116 & 572 & 210 \\
    \bottomrule
  \end{tabular} }
\end{table}

\begin{table*}[t]
\centering
\renewcommand\arraystretch{1.0}
  \caption{Experimental results of \baby on three prevalent MD datasets \textit{GossipCop}, \textit{PolitiFact} and \textit{Snopes}. The results indicated by * represent that they are statistically significant than their compared baseline models (p-value < 0.05).}
  \label{result}
  \setlength{\tabcolsep}{5pt}{
  \begin{tabular}{m{3.30cm}m{1.45cm}<{\centering}m{1.45cm}<{\centering}m{1.45cm}<{\centering}m{1.45cm}<{\centering}m{1.45cm}<{\centering}m{1.45cm}<{\centering}m{1.45cm}<{\centering}}
    \toprule
    \quad \quad \quad \quad Method & Macro F1 & Accuracy & Precision & Recall & F1$_\text{real}$ & F1$_\text{fake}$ & AUC \\
    \hline
    \multicolumn{8}{c}{\textbf{Dataset: \textit{GossipCop}}} \\
    BERT$_{\text{base}}$ \citep{devlin2019bert} ($\sim$\textit{110M}) & 78.23{\small \color{grayv} $\pm$0.45} & 83.78{\small \color{grayv} $\pm$0.80} & 79.00{\small \color{grayv} $\pm$1.45} & 77.69{\small \color{grayv} $\pm$0.59} & 89.21{\small \color{grayv} $\pm$0.69} & 67.24{\small \color{grayv} $\pm$0.45} & 86.58{\small \color{grayv} $\pm$0.33} \\
    BERT + EANN \citep{wang2018eann} & 78.59{\small \color{grayv} $\pm$0.84} & 84.47{\small \color{grayv} $\pm$0.66} & 80.37{\small \color{grayv} $\pm$1.46} & 77.42{\small \color{grayv} $\pm$1.36} & 89.80{\small \color{grayv} $\pm$0.55} & 67.39{\small \color{grayv} $\pm$1.59} & 86.89{\small \color{grayv} $\pm$0.45} \\
    BERT + BERT-EMO \citep{zhang2021mining} & 78.63{\small \color{grayv} $\pm$0.47} & 84.62{\small \color{grayv} $\pm$0.39} & 79.75{\small \color{grayv} $\pm$0.93} & 77.10{\small \color{grayv} $\pm$1.01} & 89.83{\small \color{grayv} $\pm$0.59} & 67.23{\small \color{grayv} $\pm$1.03} & 86.75{\small \color{grayv} $\pm$0.37} \\
    BERT + CED \citep{wu2023category} & 78.33{\small \color{grayv} $\pm$0.40} & 83.77{\small \color{grayv} $\pm$0.68} & 78.85{\small \color{grayv} $\pm$1.26} & 77.94{\small \color{grayv} $\pm$0.25} & 89.17{\small \color{grayv} $\pm$0.57} & 67.49{\small \color{grayv} $\pm$0.25} & 86.31{\small \color{grayv} $\pm$0.46} \\
    \hdashline
    T5$_{\text{base}}$ \citep{raffel2020exploring} ($\sim$\textit{220M}) & 78.44{\small \color{grayv} $\pm$0.33} & 84.56{\small \color{grayv} $\pm$0.27} & 80.61{\small \color{grayv} $\pm$0.50} & 76.92{\small \color{grayv} $\pm$0.33} & 89.93{\small \color{grayv} $\pm$0.20} & 66.96{\small \color{grayv} $\pm$0.51} & 87.56{\small \color{grayv} $\pm$0.36} \\
    \rowcolor{lightgrayv} \quad + \baby (ours) & 79.45{\small \color{grayv} $\pm$0.62}* & 85.33{\small \color{grayv} $\pm$0.49}* & 81.92{\small \color{grayv} $\pm$1.08}* & 77.82{\small \color{grayv} $\pm$0.88}* & 90.44{\small \color{grayv} $\pm$0.39} & 68.46{\small \color{grayv} $\pm$1.12}* & 87.46{\small \color{grayv} $\pm$0.12} \\
    T5 + EANN \citep{wang2018eann} & 78.60{\small \color{grayv} $\pm$0.35} & 84.40{\small \color{grayv} $\pm$0.18} & 80.06{\small \color{grayv} $\pm$0.51} & 77.51{\small \color{grayv} $\pm$0.74} & 89.74{\small \color{grayv} $\pm$0.18} & 67.47{\small \color{grayv} $\pm$0.77} & 87.49{\small \color{grayv} $\pm$0.54} \\
    \rowcolor{lightgrayv} \quad + \baby (ours) & 79.73{\small \color{grayv} $\pm$0.66}* & 85.59{\small \color{grayv} $\pm$0.48}* & 82.38{\small \color{grayv} $\pm$0.91}* & 77.96{\small \color{grayv} $\pm$0.85} & 90.63{\small \color{grayv} $\pm$0.66}* & 68.83{\small \color{grayv} $\pm$1.00}* & 87.97{\small \color{grayv} $\pm$0.22} \\
    T5 + BERT-EMO \citep{zhang2021mining} & 78.46{\small \color{grayv} $\pm$0.39} & 84.48{\small \color{grayv} $\pm$0.26} & 80.45{\small \color{grayv} $\pm$0.89} & 77.11{\small \color{grayv} $\pm$1.08} & 89.84{\small \color{grayv} $\pm$0.31} & 67.08{\small \color{grayv} $\pm$1.01} & 87.53{\small \color{grayv} $\pm$0.36} \\
    \rowcolor{lightgrayv} \quad + \baby (ours) & 79.55{\small \color{grayv} $\pm$0.53}* & 85.19{\small \color{grayv} $\pm$0.26}* & 81.34{\small \color{grayv} $\pm$0.63}* & 78.26{\small \color{grayv} $\pm$1.01}* & 90.29{\small \color{grayv} $\pm$0.21} & 68.81{\small \color{grayv} $\pm$1.07}* & 88.10{\small \color{grayv} $\pm$0.15}* \\
    T5 + CED \citep{wu2023category} & 78.83{\small \color{grayv} $\pm$0.64} & 84.49{\small \color{grayv} $\pm$0.52} & 80.09{\small \color{grayv} $\pm$0.89} & 77.86{\small \color{grayv} $\pm$0.82} & 89.77{\small \color{grayv} $\pm$0.39} & 67.90{\small \color{grayv} $\pm$1.04} & 87.57{\small \color{grayv} $\pm$0.36} \\
    \rowcolor{lightgrayv} \quad + \baby (ours) & 79.78{\small \color{grayv} $\pm$0.60}* & 85.50{\small \color{grayv} $\pm$0.38}* & 82.01{\small \color{grayv} $\pm$0.86}* & 78.24{\small \color{grayv} $\pm$0.91} & 90.53{\small \color{grayv} $\pm$0.27}* & 69.03{\small \color{grayv} $\pm$1.08}* & 88.01{\small \color{grayv} $\pm$0.47} \\
    \hline
    \specialrule{0em}{0.5pt}{0.5pt}
    \hline
    \multicolumn{8}{c}{\textbf{Dataset: \textit{PolitiFact}}} \\
    BERT$_{\text{base}}$ \citep{devlin2019bert} ($\sim$\textit{110M}) & 59.46{\small \color{grayv} $\pm$0.98} & 60.02{\small \color{grayv} $\pm$0.73} & 60.26{\small \color{grayv} $\pm$0.91} & 59.82{\small \color{grayv} $\pm$0.88} & 62.38{\small \color{grayv} $\pm$1.58} & 59.55{\small \color{grayv} $\pm$1.02} & 64.28{\small \color{grayv} $\pm$1.22} \\
    BERT + CED \citep{wu2023category} & 60.11{\small \color{grayv} $\pm$0.59} & 60.33{\small \color{grayv} $\pm$0.85} & 60.55{\small \color{grayv} $\pm$0.93} & 60.35{\small \color{grayv} $\pm$0.70} & 61.08{\small \color{grayv} $\pm$1.41} & 59.16{\small \color{grayv} $\pm$1.84} & 64.71{\small \color{grayv} $\pm$0.88} \\
    \hdashline
    T5$_{\text{base}}$ \citep{raffel2020exploring} ($\sim$\textit{220M}) & 59.09{\small \color{grayv} $\pm$1.32} & 59.53{\small \color{grayv} $\pm$0.98} & 59.71{\small \color{grayv} $\pm$0.92} & 59.42{\small \color{grayv} $\pm$1.13} & 61.23{\small \color{grayv} $\pm$1.70} & 56.95{\small \color{grayv} $\pm$1.83} & 63.81{\small \color{grayv} $\pm$1.14} \\
    \rowcolor{lightgrayv} \quad + \baby (ours) & 60.31{\small \color{grayv} $\pm$0.89}* & 60.67{\small \color{grayv} $\pm$0.78}* & 60.72{\small \color{grayv} $\pm$0.86}* & 60.47{\small \color{grayv} $\pm$0.86}* & 63.22{\small \color{grayv} $\pm$1.40}* & 57.40{\small \color{grayv} $\pm$1.31} & 64.98{\small \color{grayv} $\pm$1.10}* \\
    
    T5 + CED \citep{wu2023category} & 59.19{\small \color{grayv} $\pm$0.97} & 59.43{\small \color{grayv} $\pm$0.89} & 59.39{\small \color{grayv} $\pm$0.97} & 59.28{\small \color{grayv} $\pm$0.89} & 61.70{\small \color{grayv} $\pm$1.37} & 56.69{\small \color{grayv} $\pm$1.27} & 63.63{\small \color{grayv} $\pm$0.85} \\
    \rowcolor{lightgrayv} \quad + \baby (ours) & 61.27{\small \color{grayv} $\pm$1.11}* & 61.42{\small \color{grayv} $\pm$0.93}* & 61.41{\small \color{grayv} $\pm$0.82}* & 61.33{\small \color{grayv} $\pm$0.74}* & 63.09{\small \color{grayv} $\pm$1.43}* & 59.45{\small \color{grayv} $\pm$0.71}* & 65.73{\small \color{grayv} $\pm$1.16}* \\
    \hline
    \specialrule{0em}{0.5pt}{0.5pt}
    \hline
    
    \multicolumn{8}{c}{\textbf{Dataset: \textit{Snopes}}} \\
    BERT$_{\text{base}}$ \citep{devlin2019bert} ($\sim$\textit{110M}) & 62.28{\small \color{grayv} $\pm$1.21} & 71.55{\small \color{grayv} $\pm$1.57} & 63.27{\small \color{grayv} $\pm$1.32} & 62.05{\small \color{grayv} $\pm$1.22} & 43.67{\small \color{grayv} $\pm$1.66} & 80.89{\small \color{grayv} $\pm$1.61} & 69.48{\small \color{grayv} $\pm$1.32} \\
    BERT + CED \citep{wu2023category} & 62.68{\small \color{grayv} $\pm$0.78} & 71.91{\small \color{grayv} $\pm$1.44} & 63.59{\small \color{grayv} $\pm$1.28} & 62.29{\small \color{grayv} $\pm$0.94} & 44.17{\small \color{grayv} $\pm$0.97} & 81.20{\small \color{grayv} $\pm$1.41} & 70.42{\small \color{grayv} $\pm$0.74} \\
    \hdashline
    T5$_{\text{base}}$ \citep{raffel2020exploring} ($\sim$\textit{220M}) & 62.51{\small \color{grayv} $\pm$0.91} & 72.19{\small \color{grayv} $\pm$1.26} & 63.73{\small \color{grayv} $\pm$1.11} & 62.03{\small \color{grayv} $\pm$1.10} & 43.51{\small \color{grayv} $\pm$1.44} & 81.52{\small \color{grayv} $\pm$1.26} & 70.70{\small \color{grayv} $\pm$0.27} \\
    \rowcolor{lightgrayv} \quad + \baby (ours) & 64.21{\small \color{grayv} $\pm$0.82}* & 73.85{\small \color{grayv} $\pm$1.54}* & 66.08{\small \color{grayv} $\pm$0.95}* & 63.56{\small \color{grayv} $\pm$1.17}* & 45.71{\small \color{grayv} $\pm$1.53}* & 82.72{\small \color{grayv} $\pm$1.04}* & 70.59{\small \color{grayv} $\pm$0.59} \\
    
    T5 + CED \citep{wu2023category} & 62.70{\small \color{grayv} $\pm$0.28} & 72.32{\small \color{grayv} $\pm$1.43} & 63.23{\small \color{grayv} $\pm$1.00} & 62.52{\small \color{grayv} $\pm$0.45} & 44.82{\small \color{grayv} $\pm$1.31} & 80.58{\small \color{grayv} $\pm$1.47} & 69.15{\small \color{grayv} $\pm$0.90} \\
    \rowcolor{lightgrayv} \quad + \baby (ours) & 64.51{\small \color{grayv} $\pm$1.05}* & 74.14{\small \color{grayv} $\pm$0.66}* & 66.21{\small \color{grayv} $\pm$1.72}* & 63.80{\small \color{grayv} $\pm$0.80}* & 46.35{\small \color{grayv} $\pm$1.04}* & 82.67{\small \color{grayv} $\pm$1.05}* & 70.44{\small \color{grayv} $\pm$1.08}* \\
    \bottomrule
  \end{tabular} }
\end{table*}

\begin{table}[t]
\centering
\renewcommand\arraystretch{1.10}
  \caption{Ablative study. The bold and underlined scores indicate the highest and lowest results in the ablative versions.}
  \label{ablative}
  \small
  \setlength{\tabcolsep}{5pt}{
  \begin{tabular}{m{1.80cm}m{0.60cm}<{\centering}m{0.65cm}<{\centering}m{0.60cm}<{\centering}m{0.60cm}<{\centering}m{0.65cm}<{\centering}m{0.65cm}<{\centering}}
    \toprule
    \quad Method & F1 & Acc. & P. & R. & F1$_\text{real}$ & F1$_\text{fake}$ \\
    \hline
    \multicolumn{7}{c}{\textbf{Dataset: \textit{GossipCop}}} \\
    T5$_{\text{base}}$ \citep{raffel2020exploring} & 78.44 & 84.56 & 80.61 & 76.92 & 89.93 & 66.96 \\
    \rowcolor{lightgrayv} \quad + \baby & 79.45 & 85.33 & 81.92 & 77.82 & 90.44 & 68.46 \\
    \hdashline
    w/o $\mathcal{L}_D$ & 78.80 & 84.56 & 80.34 & 77.65 & 89.85 & 67.75 \\
    w/o hierarchy & 78.93 & 84.68 & 80.65 & 77.70 & 89.91 & 67.92 \\
    w direct query & \underline{78.01} & \underline{84.22} & \underline{80.12} & \underline{76.56} & \underline{89.69} & \underline{66.32} \\
    w/o weights & \textbf{79.21} & \textbf{84.82} & \textbf{80.68} & \textbf{78.17} & \textbf{90.00} & \textbf{68.43} \\
    
    \hline
    \specialrule{0em}{0.5pt}{0.5pt}
    \hline
    \multicolumn{7}{c}{\textbf{Dataset: \textit{PolitiFact}}} \\
    T5$_{\text{base}}$ \citep{raffel2020exploring} & 59.09 & 59.53 & 59.71 & 59.42 & 61.23 & 56.95 \\
    \rowcolor{lightgrayv} \quad + \baby & 60.31 & 60.67 & 60.72 & 60.47 & 63.22 & 57.40 \\
    \hdashline
    w/o $\mathcal{L}_D$ & 59.63 & 60.43 & 60.53 & 60.05 & \textbf{64.23} & \underline{55.03} \\
    w/o hierarchy & 59.73 & 60.11 & 60.31 & 59.95 & 61.59 & \textbf{57.84} \\
    w direct query & \underline{59.16} & \underline{59.59} & \underline{59.86} & \underline{59.51} & \underline{60.97} & 57.34 \\
    w/o weights & \textbf{60.10} & \textbf{60.83} & \textbf{61.19} & \textbf{60.58} & 63.40 & 56.81 \\
    
    \hline
    \specialrule{0em}{0.5pt}{0.5pt}
    \hline
    \multicolumn{7}{c}{\textbf{Dataset: \textit{Snopes}}} \\
    T5$_{\text{base}}$ \citep{raffel2020exploring} & 62.51 & 72.19 & 63.73 & 62.03 & 43.51 & 81.52 \\
    \rowcolor{lightgrayv} \quad + \baby & 64.21 & 73.85 & 66.08 & 63.56 & 45.71 & 82.72 \\
    \hdashline
    w/o $\mathcal{L}_D$ & 63.63 & \textbf{74.11} & \textbf{66.00} & 62.74 & 44.12 & \textbf{83.13} \\
    w/o hierarchy & 63.86 & 72.97 & 64.79 & 63.30 & \textbf{45.76} & 81.98 \\
    w direct query & \underline{62.68} & \underline{72.09} & \underline{63.61} & \underline{62.20} & \underline{43.96} & \underline{81.39} \\
    w/o weights & \textbf{63.96} & 73.37 & 65.41 & \textbf{63.38} & 45.61 & 82.33 \\
    \bottomrule
  \end{tabular} }
\end{table}

\vspace{3pt}
\noindent
\textbf{Implementation details}.
In the experiments conducted in this work, we employ a pre-trained version \textit{bert-base-uncased} \citep{devlin2019bert}\footnote{Downloaded from \url{https://huggingface.co/bert-base-uncased}.} as the backbone for the BERT-based baselines, and initialize the T5-based models using \textit{flan-t5-base} \citep{chung2022scaling}\footnote{Downloaded from \url{https://huggingface.co/t5-base}.}. To mitigate overfitting, we empirically freeze the first 10 Transformer layers \citep{vaswani2017attention} of BERT, and allow only the final layer to be trained; for the T5 model, we also train only the last layer of both the encoder and decoder. Our feature fusion network utilizes a two-head attention network, and the veracity classifier employs a standard feed-forward neural network, which stacks two linear layers and two ReLU non-linear functions. The optimization process of model parameters involves an Adam optimizer \citep{kingma2015adam}, with a fixed learning rate set to $7 \times 10^{-5}$ for both the pre-trained model and other modules, and maintains a constant batch size of 64. Additionally, an early stopping strategy is implemented, which stops training if the model does not exhibit better macro F1 values on the validation set for 10 consecutive epochs. The hyper-parameter $\beta$ is empirically fixed to $10^{-4}$.

\subsection{Main Results (Q1)}

We measure our \baby with 7 typical evaluation metrics in the experiments, including accuracy, precision, recall, macro F1 value, F1 values F1$_\text{real}$ and F1$_\text{fake}$ corresponding to two categories, and Area Under Curve (AUC). Table~\ref{result} reports the main experimental results, which presents the performance of compared baselines and their improved versions with \baby. In general, this paper proposes a plug-and-play method that consistently enhances the performance of its baseline models across almost all settings. For instance, on the \textit{GossipCop} dataset, utilizing EANN as the baseline model, our approach outperforms it by 2.32 in terms of precision; and on the \textit{Snopes} dataset, employing CED as the baseline model, \baby surpasses it by 2.98 in precision. Analytically, the basic difference between \baby and the baselines lies in the decoder-side inputs, which are intent reasoning sequences $\mathbf{z}$ and original input articles $\mathbf{x}$, respectively. Therefore, these experimental results can illustrate the significant contribution of the intent sequences generated by \baby in improving the overall performance.

\begin{figure*}[t]
  \centering
  \includegraphics[scale=0.67]{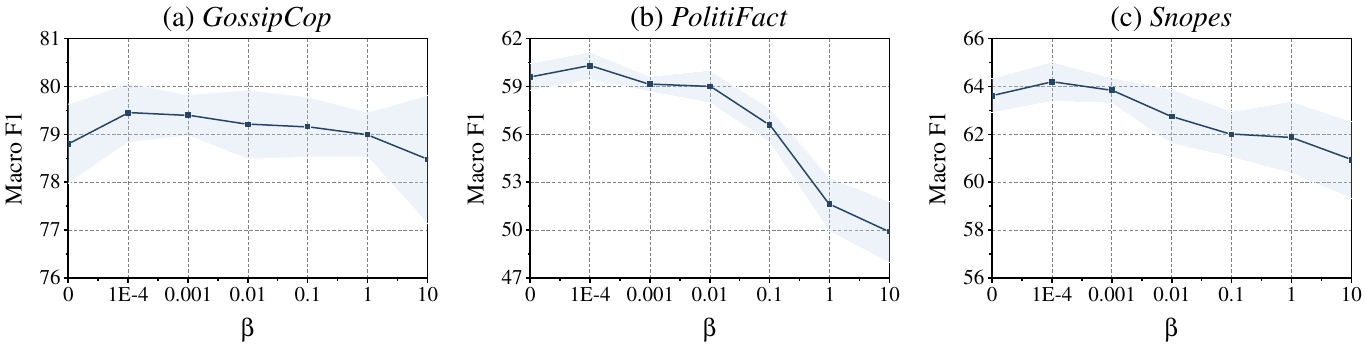}
  \caption{Sensitivity analysis of the trade-off parameter $\beta$.}
  \label{sensitivity}
\end{figure*}

Then, we proceed to analyze the performance differences of \baby across three distinct MD datasets. Generally, the ranking of the average improvement over the baseline models is \textit{Snopes} \textgreater \textit{PolitiFact} \textgreater \textit{GossipCop}. Statistically, concerning the macro F1 metric, our \baby exhibits an average improvement of 1.04, 1.65, and 1.75 on the three datasets, respectively. This phenomenon indicates that the positive impact of \baby is more pronounced in scenarios with limited training data, such as \textit{Snopes}. In other words, it highlights the ability of \baby to enhance misinformation detectors by accurately reasoning the intent behind the articles, especially in situations with smaller data scales.
Finally, we provide an analysis of the performance differences among different baseline MD models. We compare the basic T5 model with the state-of-the-art CED model and find that \baby achieves a greater improvement under the stronger CED model. For example, on the \textit{PolitiFact} dataset, the improvement brought by \baby over CED exceeds that over T5 by approximately 0.86, which can highlight the effectiveness of our plug-and-play approach in unlocking the potential of the baseline MD models. Additionally, when comparing the two backbone models, BERT and T5, we know that BERT is pre-trained with a masked language model objective, and it is objectively better suited for feature extraction and natural language understanding tasks. Even though our model uses T5 as the backbone, its performance still significantly outperforms models based on BERT.

\begin{table*}[t]
\centering
\renewcommand\arraystretch{0.95}
  \caption{Case study. We provide three cases from MD datasets to demonstrate generated reasoning results by \baby.}
  \label{case}
  \small
  \setlength{\tabcolsep}{5pt}{
  \begin{tabular}{m{5.5cm}|m{5.5cm}|m{5.5cm}}
    \bottomrule
    \textbf{Article}: {\fontsize{7.5pt}{0pt}\selectfont Warning : This article contains spoilers! So many spoilers! Highly detailed, movie-ruining spoilers! ``Somewhere out there, there’ s an 8-year-old girl dreaming of becoming a criminal,'' Debbie Ocean, played by Sandra Bullock, tells her mirrored reflection in one of the standout moments of “ Ocean’ s 8, ” the highly anticipated sequel to Steven Soderbergh’ s iconic heist films. ``You’ re doing this for her.'' The film gives budding bad girls everywhere role models to look up to, but just how...}
    & \textbf{Article}: {\fontsize{7.5pt}{0pt}\selectfont The Coachella Valley Music and Arts Festival has announced the dates for its 20th anniversary and how to get pre-sale tickets. The festival is also offering a new upgrade for car camping. So when is Coachella 2019? The festival happens April 12-14 and April 19-21 at the Empire Polo Club in Indio. As it has done in recent years, promoter Goldenvoice will put a limited number of passes on sale early. This year you can get Coachella 2019 passes for Weekend 1 and 2 starting at 11 a.m. Pacific Friday, June 1.}
    & \textbf{Article}: {\fontsize{7.5pt}{0pt}\selectfont Alex Rodriguez flatly denied a report that claimed he was threatening to cut child-support payments for his two daughters over a legal dispute with his ex-wife’ s brother. ``I have always paid far more than the maximum in child support and that will never change,'' the former New York Yankees star said in a statement to Page Six. ``It’ s highly offensive to me that my former brother-in-law, who has been trying to pursue a frivolous case against me for four years and has gotten absolutely nowhere with it ...} \\
    \rowcolor{lightgrayv} \textbf{Veracity Label}: \textit{fake} \quad \textbf{Prediction}: \textit{fake} 
    & \textbf{Veracity Label}: \textit{real} \quad \textbf{Prediction}: \textit{real} 
    & \textbf{Veracity Label}: \textit{fake} \quad \textbf{Prediction}: \textit{fake} \\
    \hdashline
    \textbf{Reasoning sequence}: & \textbf{Reasoning sequence}: & \textbf{Reasoning sequence}:  \\
     {\color{gray} \footnotesize [\texttt{Public}]} Is this article aimed at the public? {\color{bluev} \textbf{yes}}
     & {\color{gray} \footnotesize [\texttt{Public}]} Is this article aimed at the public? {\color{bluev} \textbf{yes}}
     & {\color{gray} \footnotesize [\texttt{Public}]} Is this article aimed at the public? {\color{redv} \textbf{no}} \\
     {\color{gray} \footnotesize [\texttt{Emotion}]} Is there any emotional expression in this article? {\color{bluev} \textbf{yes}}
     & {\color{gray} \footnotesize [\texttt{Emotion}]} Is there any emotional expression in this article? {\color{redv} \textbf{no}} 
     & {\color{gray} \footnotesize [\texttt{Emotion}]} Is there any emotional expression in this article? {\color{redv} \textbf{no}} \\
     {\color{gray} \footnotesize [\texttt{Individual}]} Does this article express any personal points? {\color{redv} \textbf{no}}
     & {\color{gray} \footnotesize [\texttt{Individual}]} Does this article express any personal points? {\color{redv} \textbf{no}} 
     & {\color{gray} \footnotesize [\texttt{Individual}]} Does this article express any personal points? {\color{redv} \textbf{no}} \\
    {\color{gray} \footnotesize [\texttt{Popularize}]} Is this an article aimed at popularization? {\color{redv} \textbf{no}}
     & {\color{gray} \footnotesize [\texttt{Popularize}]} Is this an article aimed at popularization? {\color{bluev} \textbf{yes}} 
     & {\color{gray} \footnotesize [\texttt{Nointent}]} This article does not convey any intents. \\
    {\color{gray} \footnotesize [\texttt{Clout}]} Is this an article aimed at pursuing attention? {\color{bluev} \textbf{yes}}
     & {\color{gray} \footnotesize [\texttt{Clout}]} Is this an article aimed at pursuing attention? {\color{redv} \textbf{no}} 
     &  \\
     {\color{gray} \footnotesize [\texttt{Conflict}]} Is this article attempting to create conflict? {\color{redv} \textbf{no}}
     & 
     &  \\
    \bottomrule
  \end{tabular} }
\end{table*}

\subsection{Ablative Study (Q2)}

To empirically answer Q2, we conduct ablative experiments to assess the positive effect of each key component in \baby. The experiments are implemented using T5 as the baseline model across all three MD datasets, and the results are shown in Table~\ref{ablative}. Our used ablative versions of \baby are briefly described as follows:

\begin{itemize}
    \item \textbf{w/o} $\mathcal{L}_D$ indicates the version without optimizing the model parameters using $\mathcal{L}_D$ in Eq.~\eqref{eq3}, equivalent to setting the parameter $\beta$ to 0;
    \item \textbf{w/o hierarchy} indicates the version without utilizing the intent hierarchy designed in this paper for intent reasoning. In other words, it flattens the intent hierarchy in Fig.~\ref{flow}, and inputs all intent queries $\mathcal{Q}$ into the model during reasoning;
    \item \textbf{w direct query} indicates the version without the intent reasoning process, instead, it directly queries the model at the decoder side, ``\textit{what is the intent behind this article?}'' and outputs a short answer;
    \item \textbf{w/o weights} indicates the version without employing the weights proposed in this paper for sample filtering.
\end{itemize}

In general, the performance of all ablation experiments is consistently lower than that of our comprehensive model \baby, providing a positive response to Q2, as each module designed in our model contributes positively. According to statistical results, the performance of the four ablation versions can be ranked as: w direct query \textless\ w/o $\mathcal{L}_D$ \textless\ w/o hierarchy \textless\ w/o weights. In more detail, the version with direct query exhibits the poorest performance, even falling below the baseline T5 model on the \textit{GossipCop} dataset, so that \baby has a negative effect on the baseline model. Analytically, this version feeds the identical query to the decoder for each sample, and yields answers consisting of only one or two words. This will result in minimal differences in intent features $\mathbf{\widehat h}^d$ between different samples, subsequently reducing the discriminative nature of the overall feature $\mathbf{e}$ through the fusion network. Furthermore, directly prompting large models for intent answers may lead to entirely incorrect intent reasoning, introducing significant noise into the model.

The version without $\mathcal{L}_D$ also exhibits inferior performance compared to other ablation versions, which highlights the importance of $\mathcal{L}_D$ and indicates that supervising the model with a self-supervised objective function enables accurate binary classification results even on unseen samples. The version without hierarchy also experiences a decline compared to our complete model \baby, which can demonstrate the beneficial role of the reasoning process in accurately predicting intents. Additionally, if we employ $\mathcal{Q}$ to infer all intents, the input sequence at the decoder side may be so long and non-discriminative that affects the performance of models, and the hierarchical structure proposed in this paper effectively addresses this issue. Finally, the assigned weights also play a role, although its impact is not as pronounced as other ablation results. We argue this partly reflects that errors in reasoning intents by large language models are relatively scarce, which demonstrates the potential of these large models and the success of formulating intent classification into the binary classification task.

\subsection{Sensitivity Analysis (Q2)}

To further investigate Q2, we conduct a sensitivity analysis on a crucial hyper-parameter, $\beta$ in Eq.~\eqref{eq3}, to examine the model's sensitivity to $\beta$ and determine its most beneficial value. The experimental results are illustrated in Fig.~\ref{sensitivity}, where we evaluate our experiments using macro F1 on all three datasets \textit{GossipCop}, \textit{PolitiFact}, and \textit{Snopes}, and we select $\beta$ from the set $\{0, 0.0001, 0.001, 0.01, 0.1, 1, 10\}$. Generally, the model consistently shows the best performance on all datasets when $\beta$ is approximately 0.0001 and exhibits a declining trend as it decreases or increases. This suggests that without using $\mathcal{L}_D$ for training, \baby may provide inaccurate results when reasoning new intents, thereby negatively affecting the results of MD models; and as it increases, parameter optimization prioritizes intent reasoning while neglecting the objective of veracity classification in Eq.~\eqref{eq3}. Therefore, based on these empirical results, we set $\beta$ to 0.0001 for all experiments conducted in this paper.

In addition, an anomalous phenomenon is observed in Fig.~\ref{sensitivity}, especially on small-scale MD datasets such as \textit{PolitiFact}. When $\beta$ takes a higher value, the results of \baby exhibit a significant decline. For instance, with $\beta$ set to 10, the macro F1 value of the model drops to 49.90, representing a decrease of 10.41 compared to the result when $\beta$ is fixed to 0.0001. We attribute this phenomenon to two reasons: (1) $\mathcal{L}_D$, as a generative objective function, empirically requires more computational resources than discriminative tasks. On small-scale datasets like \textit{PolitiFact}, with only 2568 training samples, the training of $\mathcal{L}_D$ is insufficient and may result in \baby reasoning numerous noisy intents. (2) When $\beta$ assumes a higher value, the model's focus on the objective function for veracity classification in Eq.~\eqref{eq3} significantly diminishes during training, then exacerbating the performance of inadequately trained models on small-scale data.

\subsection{Case Study (Q3)}

In this section, we provide illustrative examples to partially answer Q3. We select three representative cases from the \textit{GossipCop} dataset and show them in Table~\ref{case}. Specifically, the first example employs some exaggerated vocabulary, \eg ``Warning,'' to capture the reader's attention, and \baby accurately identifies the intent of the article as \texttt{Clout}. The second example primarily popularizes information about the Coachella music festival's schedule, and \baby also provides accurate answers. In the final case, the article simply shares information without manifesting a clear intent, leading \baby to respond no to all queries. In summary, these cases consistently demonstrate the effectiveness of \baby in reasoning intents.

\section{Conclusion and Future Work}

In this work, we present to investigate the intents expressed by articles and utilize them to identity misinformation. Therefore to achieve this target, we design a new MD model named \baby, which explicitly reasons intents and captures the intent features. To be specific, we design an intent hierarchy based on several psychological studies and use it to progressively reason intents with a pre-trained auto-regressive decoder. Additionally, we also assign adaptive weights for training samples to avoid error propagation and veracity inconsistency issues. Our experimental results can indicate that \baby can improve the performance of the baseline models, and we also provide some representative cases to demonstrate that \baby can accurately reason intents. Furthermore, we also declare a limitation existing in our work, which can be improved in our future work. We construct the intent hierarchy by collecting different types of intents and organizing them into a hierarchical structure. The hierarchy used in this article has only two layers. In the future, with deep research in psychology and sociology, we look forward to designing more complex and diverse hierarchical structures to further improve the accuracy of intent recognition.

\section*{Acknowledgements}

This work is supported by the National Science and Technology Major Project of China (No. 2021ZD0112501), the National Natural Science Foundation of China (No. 62376106, No. 62276113), and China Postdoctoral Science Foundation (No. 2022M721321).

\clearpage

\bibliographystyle{ACM-Reference-Format}
\balance
\bibliography{reference}

\end{document}